# Combining GCN and Transformer for Chinese Grammatical Error Detection

Jinhong Zhang[a, *]

[a]School of Computer Science and Cyber Engineering, Guangzhou University, Guangzhou, China, 0000-0001-9492-2003

## ABSTRACT

This paper describes our system at a task: Chinese Grammatical Error Diagnosis (CGED). The task is held by the Natural Language Processing Techniques for Educational Applications (NLP-TEA) to encourage the development of automatic grammatical error diagnosis in Chinese learning since 2014. The goal of CGED is to diagnose four types of grammatical errors: word selection (S), redundant words (R), missing words (M), and disordered words (W). The automatic CGED system contains two parts including error detection and error correction and our system is designed to solve the error detection problem. Our system is built on three models: 1) a BERT-based model leveraging syntactic information; 2) a BERT-based model leveraging contextual embeddings; 3) a lexicon-based graph neural network leveraging lexical information. We also design an ensemble mechanism to improve the single model's performance. Finally, our system achieves the highest F1 scores at detection level and identification level among all teams participating in the CGED 2020 task.

## 1. Introduction

The Chinese language is often recognized as one of the most difficult to learn. In comparison to English, Chinese does not have a singular/plural transition, nor does it have verb tense variations. Furthermore, because word boundaries are not explicitly specified in Chinese, word segmentation is frequently required prior to deeper analysis. All these problems make Chinese learning challenging to new learners. In recent years, more and more people with different language and knowledge background have become interested in learning Chinese as a second language. To assist in identifying and correcting grammatical errors produced by these people, an automatic Chinese Grammatical Error Diagnosis (CGED) tool is required.

Since 2014, the Natural Language Processing Techniques for Educational Applications (NLP-TEA) have taken CGED as one of the shared tasks to encourage the development of automatic grammatical error diagnosis in Chinese learning. Many methods have been proposed to solve CGED task.

In this work, we introduce our system to solve the error detection problem. In our system, we use three types of models. The first one is the BERT-GCN-LSTM-CRF, which is based on the model of multi-layer bidirectional transformer encoder and incorporates GCN to improve the performance. The second one is the BERT with context-LSTM-CRF, which makes use of contextualized word representations because they have the ability to efficiently capture compositional information in language. The third one is the LGN, which incorporates lexical information into Chinese NER tasks.

We also design an ensemble mechanism to improve the single model's performance. In the experiment, our system gets the highest F1 scores at detection level and identification level among all the models that participated in the NLPTEA-2020 CGED task.

## 2. Chinese Grammatical Error Diagnosis

Since 2014, the CGED shared task has been held. Several sets of training data written by CFL learners that contain numerous grammatical errors have been released. For detection, the CGED defines four types of errors: (1) R (redundant word errors); (2) M (missing words); (3) W (word ordering errors); (4) S (word selection errors). Table 1 shows some typical examples. The performance is measured at three levels: detection, identification, and position. For correction, systems need to recommend no more than three corrections for missing and selection errors. In this paper, our system focuses on the error detection problem.

## 3. Models

Some previous works consider the error detection problem to be a sequence labeling problem. Similarly, we use BIO encoding to generate a corresponding label sequence $y$ for a sentence $x$ [1]. We use three models to solve the labeling problem, in which BERT and GCN information are all used.

### 3.1. BERT-GCN-LSTM-CRF

Previous works use LSTM-CRF model to solve the problem [2]. For better performance, we combine the BERT model and GCN model.

We use BERT [3], the multi-layer bidirectional transformer encoder [4] to encode the input sentence. As shown in Fig. 1 Architectures of BERT-GCN-LSTM-CRF for grammatical error detection, given an input sequence $S = x_1, x_2, \cdots, x_N$, BERT outputs the hidden states $S' = h_1, h_2, \cdots, h_N$.

* Corresponding author. e-mail: jhzhang@e.gzhu.edu.cn

### 3.1.1. GCN

Previous works [2][5] spent a lot of effort in feature engineering including pretrained features and parsing features. Part-of-speech-tagging (POS), and dependency information are the most important parsing features, which indicates to us the task is closely associated with the structure of the sentence syntactic dependency.

To understand the dependency structure of an input sentence better, we introduce the Graph Convolution Network (GCN) [6][7]. Specifically, we use the graph attention networks (GAT) [8] to assign different importance to nearby nodes using masked self-attention layers. The multi-layer GCN network accepts the high-level character information obtained by the BERT model and the adjacency matrix of the dependency tree. A GAT operation with M independent attention heads can be expressed as follows:

$$f_i' = \prod_{m=1}^{M} \sigma \left( \sum_{j \in N_i} \alpha_{ij}^m W^m f_j \right) \quad (1)$$

$$\alpha_{ij}^m = \frac{\exp\left(LeakyReLU(a^T[W^m f_i \| W^M f_j])\right)}{\sum_{m \in N_i} \exp(LeakyReLU(a^T[W^m f_i \| W^M f_m]))} \quad (2)$$

where $\prod$ is the concatenation operation, $\sigma$ is a nonlinear activation function, $N_i$ is the graph's neighborhood of node $i$, $\alpha_{ij}^m$ are the attention coefficients and $a$ is a feed-forward neural network. At the last layer, averaging will be adopted:

$$f_i^{final} = \sigma \left( \frac{1}{M} \sum_{m=1}^{M} \sum_{j \in N_i} \alpha_{ij}^m W^m f_j \right) \quad (3)$$

### 3.1.2. Concatenation

Following the graph convolution network, we concatenate the representation $H_l$ for the l-th layer with the BERT hidden state as the LSTM layer's input.

### 3.1.3. CRF

To predict the sequence tags for each token, a CRF layer is added.

$$Score(X, Y) = \sum_{i=0}^{n} A_{y_i, y_{i+1}} + \sum_{i=1}^{n} V_{i, y_i} \quad (4)$$

$$P(Y|X) = \frac{\exp(Score(X, Y))}{\sum_{Y'} \exp(Score(X, Y'))} \quad (5)$$

where X, Y, $Y'$ represents the input sequence, the truth tag sequence, and an arbitrary label sequence, V represents the emission scores, and A is the transition scores matrix of the CRF layer. We use Viterbi Decoding [2] to inference answers.

### 3.2. BERT with context-LSTM-CRF

Error detection can be difficult because CGED datasets are restricted in size and the label distributions are extremely unbalanced. As described in [9], Contextualized word representations can capture compositional information in language efficiently, and they can be optimized on large amounts of unsupervised data. Specifically, it uses ELMo, BERT and Flair embeddings as contextualized word representations. To improve the performance on this task, we similarly use the structure shown in Fig. 2 Architectures of BERT with context-LSTM-CRF for grammatical error detection.

### 3.3. LGN

RNN is frequently used in Chinese named entity recognition (NER) task. However, because of the chain structure and the lack of global semantics, RNN-based models are vulnerable to word ambiguities. LGN [10] alleviates this problem by introducing a lexicon-based graph neural network with global semantics, in which lexicon knowledge is used to connect characters to capture the local composition, while a global relay node connects each character node and word edge to capture global sentence semantics and long-range dependency. Fig. 3 Architectures of LGN for grammatical error detection shows the structure of LGN. Node c represents every character and e represents every potential word. The model can use global context information to repeatedly compare ambiguous words to tackle the word ambiguities problem based on the multiple graph-based interactions among characters, potential words, and the whole-sentence semantics. LGN achieves Chinese NER as a graph node classification task. We treat error detection task as NER and use LGN to increase the diversity of prediction.

## 4. Ensemble Mechanism

To generate better results, we train multiple error detection models, and employ a three-stage voting ensemble mechanism to get the final result by utilizing the predictions from multiple models.

Before all, we convert the BIO encoding to the format like: (error_start_position, error_end_position, error_type) and then use the ensemble mechanism:

In the first stage, we calculate the number of errors of each type. If the number of errors of a certain type is greater than $\theta_1$ of the number of all models, we believe that there is an error of this type and the error is the largest of all predictions of this type of error. Specially, we do not include LGN when calculating the number of models for the reason that the number of predictions of LGN is small.

In the second stage, if an error appears in the predictions of more than $\theta_2$ models, we think the error exits. Also, we do not include LGN when calculating the number of models if the error is not predicted by LGN for the same reason.

In the third stage, we calculate the number of all predicted errors. If it is great than $\theta_3$ of the number of all models, we believe there is an error. If no errors are predicted after the first two stages, we think the error is the one which is predicted the most by the all models. Also, we do not include LGN when calculating the number of models for the same reason.

In the experiments, we select the $\theta_1$, $\theta_2$ and $\theta_3$ according to the performance on the validation data.

## 5. Experiments

### 5.1. Data and Experiment Settings

We trained our single models using training units that include both the incorrect and the corrected sentences from 2016 (HSK Track), 2017, 2018, 2020 training data sets, as well as 2016 (HSK Track) and 2018 testing data sets. The sentences from 2017 testing data set are used for validation and 2020 testing data set are used for test. The overall data distribution in the training data is shown in Table 2.

For the BERT-GCN-LSTM-CRF model, we select ELECTRA discriminator [11] as the Bert's initialization. More concretely, we use Chinese ELECTRA-Large discriminator model with 1024 hidden units, 16 heads, 24 hidden layers, 324M parameters. For the GCN model, Language Technology Platform (LTP) [12] was used to generate the dependency tree, and the first layer's hidden vector size was 512 with 8 heads, while the second layer's hidden vector size was 1024 with 8 heads. For LSTM, the hidden size was 2048 with 1 layer. For other parameters, we use streams of 128 tokens, a mini-batch of size 32, learning rate of 2e-5 and epoch of 120. We use different random seeds and dropout [13] values to train 12 single models for the ensemble mechanism.

For the BERT with context-LSTM-CRF model, we also select ELECTRA discriminator as the Bert's initialization, and we use ROBERTA to get the contextual embeddings. Specially, contextual embeddings are not fine-tuned in all experiments. Other parameters are the same as above. Also, we use different random seeds and dropout values to train 10 single models for the ensemble mechanism.

For the LGN model, the batch size, learning rate, and epoch were set to 32, 2e-5, 120. Moreover, the dropout rates for embedding, attention and aggregation module were all set to 0.1. We use different random seeds to train 45 single models for the ensemble mechanism.

*5.2. Metric*

For the error detection task, the evaluation method includes three levels:

**Detection level**. Determine whether a sentence is correct or not. The sentence is incorrect if there is an error. All types of errors will be considered incorrect.

**Identification level**. This level could be thought of as a multi-class categorization problem. The correction situation should be identical to the gold standard for a specific type of error.

**Position level**. The system's outputs should be exactly the same as the gold standard's quadruples.

At the detection, identification, and position level, the three metrics precision, recall, and f1 are measured.

*5.3. Validation Results*

We use the three single models described above as our baseline models. The results of different models are listed in Table 3.

The first and second models have a good fit on the validation set. In addition, LGN does not perform well, but mainly because of the low recall, i.e., fewer errors in the sentences are identified, while the high precision ensures a better accuracy of the identified errors. Moreover, LGN also increases the distribution of results to improve the performance of the system. To reduce the effect of low recall, we do not fully include the LGN models when using the ensemble mechanism. As shown in Table 3, The ensembled model does not achieve much improved performance on the validation set, it is mainly because the first and second models already have a good fit on the validation set and the number of training models is not large enough.

*5.4. Testing Results*

Table 4 shows the performances on error detection. Our system achieves the best F1 scores at the detection level and identification level by a balanced precision and recall among all teams participating in the CGED 2020 task. At the detection level, we improved the F1 value by 0.47% over the state-of-the-art[25], this is because we added syntactic information of the sentences, which is much richer than the POS Score and PMI Score used by the state-of-the-art method. At the identification level, we improved the F1 value by 1.23% over the state-of-the-art[23], we think this is because the state-of-the-art method only adds ResNet on top of BERT, but we not only add rich information: syntactic information, contextual embeddings and lexical information, but also add CRF layer to improve the performance, so we can get better F1 value.

Although we achieve the highest F1 score, there still has a wide gap for our system to solve the Chinese grammatical error diagnosis.

**6. Related work**

The researchers have proposed many different technologies to study the detection and correction of English grammatical errors [14][15][16]. However, there are few studies on grammatical errors in modern Chinese. Starting in 2014, the Natural Language Processing Techniques for Educational Applications (NLPTEA) has added modern Chinese grammatical error diagnosis tasks. Many methods are proposed to solve this task [17][18]. [19] proposed a model based on stacked LSTM and CRF in 2016, which improved the accuracy and recall rate of automatic grammatical error recognition. [20] used Bi-LSTM to detect the location of errors and added additional linguistic information, POS, and n-gram, combining machine learning and traditional n-gram methods. [21] regarded the error correction task as a translation task. Errors are divided into surface errors and grammatical errors. The similar phonetic table and 5-gram language model are used to solve low-level errors, and the Transformer model based on character granularity and word granularity are used to solve high-level errors. [22] adopted a multi-model parallel structure, using three types of models: rule-based, statistics-based, and neural network. To solve the detection problem, [23] combined ResNet and BERT, and explored stepwise ensemble selection from model libraries to improve the performance of the single model. [24] leveraged syntactic information and adopted a multi-task learning framework based on BERT to improve the baseline model to detect grammatical errors.

**7. Conclusion**

The paper describes our system on NLPTEA-2020 CGED task, which combines GCN and BERT for Chinese Grammatical Error Diagnosis. We also design an ensemble mechanism to maximize the model's capability. Among all teams participating in the CGED 2020 task, we achieve the highest F1 scores at detection level and identification level.

**Acknowledgements**

This research did not receive any specific grant from funding agencies in the public, commercial, or not-for-profit sectors.

**Table 1**
Typical Error Examples.

| Error Type | Original Sentence | Correct Sentence |
|---|---|---|
| M | 在我看来，我觉得结婚是很自由的情。 | 在我看来，我觉得结婚是很自由的事情。 |
| R | 对我来说，今年的我的暑假非常特别。 | 对我来说，今年我的暑假非常特别。 |
| S | 我的多爱的画家也画抽象的画儿。 | 我的最爱的画家也画抽象的画儿。 |
| W | 我觉得应该说出真相尽可能多，但有时候人被迫说谎。 | 我觉得应该说出尽可能多真相，但有时候人被迫说谎。 |

**Table 2**
Data statistics.

|  | Error | R | M | S | W |
|---|---|---|---|---|---|
| Train | 62,661 | 13,929 | 16,672 | 27,504 | 4,556 |
| Validation | 4,871 | 1,060 | 1,269 | 2,156 | 386 |
| Test | 3,660 | 768 | 862 | 1,701 | 329 |

**Table 3**
The results of single models and ensemble model on validation dataset.

| model | Detection | | | Identification | | | Position | | |
|---|---|---|---|---|---|---|---|---|---|
|  | Precision | Recall | F1 | Precision | Recall | F1 | Precision | Recall | F1 |
| BERT-GCN-LSTM-CRF | 0.8695 | 0.8444 | 0.8568 | 0.7471 | 0.6529 | 0.6968 | 0.6111 | 0.4938 | 0.5462 |
| BERT with context-LSTM-CRF | 0.8848 | 0.8338 | 0.8586 | 0.7631 | 0.6333 | 0.6922 | 0.6305 | 0.4786 | 0.5441 |
| LGN | 0.8244 | 0.2419 | 0.3741 | 0.7313 | 0.1297 | 0.2203 | 0.4316 | 0.0631 | 0.1101 |
| Ensembled Model | 0.8633 | 0.8551 | **0.8592** | 0.7611 | 0.6698 | **0.7125** | 0.6210 | 0.5054 | **0.5572** |

**Table 4**
Error detection performances on Official Testing data sets.

| TEAM | Detection | | | Identification | | | Position | | |
|---|---|---|---|---|---|---|---|---|---|
|  | Precision | Recall | F1 | Precision | Recall | F1 | Precision | Recall | F1 |
| UNIPUS-Flaubert | 0.8782 | 0.9157 | 0.8966 | 0.6507 | 0.6420 | 0.6463 | 0.3147 | 0.2739 | 0.2929 |
| NJU-NLP | 0.8565 | 0.9757 | 0.9122 | 0.5571 | 0.8432 | 0.6709 | 0.2097 | 0.4648 | 0.2890 |
| OrangePlus | 0.9252 | 0.8600 | 0.8914 | 0.7230 | 0.6287 | 0.6726 | 0.4428 | 0.3610 | 0.3977 |
| Flying | 0.9273 | 0.6213 | 0.6736 | 0.7356 | 0.6213 | 0.6736 | 0.4320 | 0.3514 | 0.3876 |
| ours | 0.9037 | 0.9304 | **0.9169** | 0.6957 | 0.6765 | **0.6859** | 0.4185 | 0.3608 | 0.3875 |

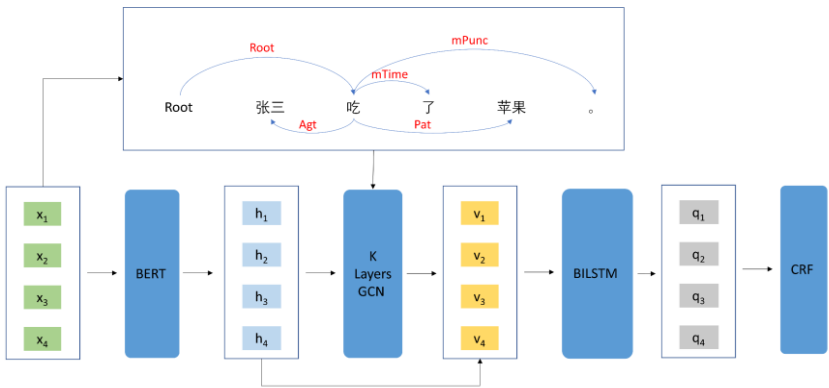

**Fig. 1** Architectures of BERT-GCN-LSTM-CRF for grammatical error detection

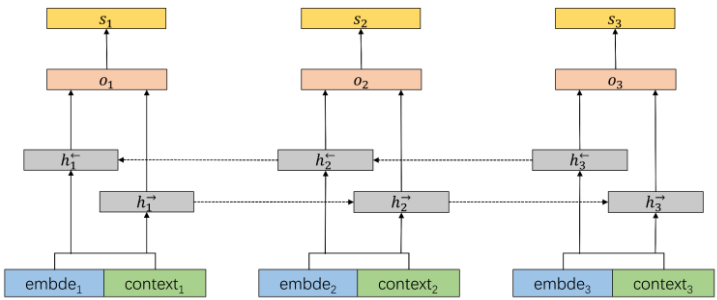

**Fig. 2** Architectures of BERT with context-LSTM-CRF for grammatical error detection

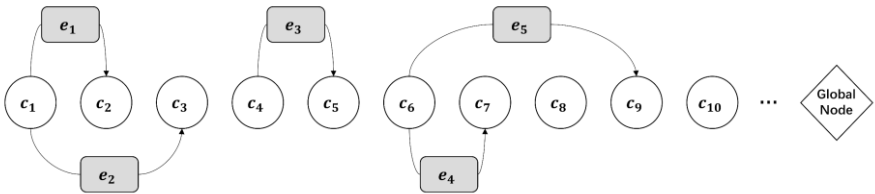

**Fig. 3** Architectures of LGN for grammatical error detection